\documentclass{article}
\usepackage{ijcai18}

\usepackage{times}
\usepackage{xcolor}
\usepackage{soul}
\usepackage[utf8]{inputenc}
\usepackage[small]{caption}

\usepackage{graphicx}
\usepackage{amsmath}
\usepackage{amssymb}

\usepackage{multirow}
\usepackage{algorithm}
\usepackage{flushend}
\usepackage[noend]{algpseudocode}

\DeclareMathOperator*{\argmin}{arg\,min}

\usepackage{subfig}

\title{Multi-Label Zero-Shot Learning with Transfer-Aware \\ Label Embedding Projection}

\author{
Meng Ye$^1$, 
Yuhong Guo$^2$, 
\\ 
$^1$ Computer and Information Sciences, Temple University, USA\\
$^2$ School of Computer Science, Carleton University, Canada\\
meng.ye@temple.edu,
yuhong.guo@carleton.ca
}

\begin{document}
\maketitle

\begin{abstract}
   Zero-shot learning 
transfers knowledge from seen classes to novel unseen classes 
to reduce human labor of labelling data for building new classifiers. 
Much effort on zero-shot learning however has focused on
the standard multi-class setting,
the more challenging multi-label zero-shot problem 
has received limited attention.
In this paper 
we propose a 
transfer-aware embedding projection approach
to tackle multi-label zero-shot learning.
The approach projects the label embedding vectors into 
a low-dimensional space to 
induce better inter-label relationships and 
explicitly facilitate information
transfer from seen labels to unseen labels,
while simultaneously learning 
a max-margin multi-label classifier with the projected label embeddings.
Auxiliary information can be conveniently incorporated to guide the label embedding projection
to further improve label relation structures for zero-shot knowledge transfer.
We conduct experiments 
for zero-shot multi-label image classification. 
The results demonstrate the efficacy of the proposed approach. 
\end{abstract}

\section{Introduction}
Despite the advances in the development of supervised learning techniques such as 
deep neural network models, the conventional supervised learning setting
requires a large number of labelled instances 
for each single class to perform training, and hence induce substantial annotation costs. 
It is important to develop algorithms that enable the reduction
of annotation cost for training classification models.
Zero-shot learning (ZSL) which 
transfers knowledge from annotated \emph{seen} classes to 
predict \emph{unseen} classes that have no labeled data, 
hence has received a lot of attention
\cite{lampert2009learning,akata2015evaluation,romera2015embarrassingly,zhang2015zero,changpinyo2017predicting}.

One primary source deployed in zero-shot learning for bridging the gap between seen and unseen classes
is the {\em attribute} description of the class labels~\cite{lampert2009learning,lampert2014attribute,romera2015embarrassingly,fu2015multiview}.
The attributes are typically defined by domain experts 
who are familiar with the common and specific characteristics of different category concepts,
and hence are able to carry transferable information across classes. 
Nevertheless 
human labor is still involved in defining the attribute-based class representations.
This propels  
the research community to exploit more easily accessible free information sources from the Internet, 
including
textual descriptions from Wikipedia articles 
\cite{qiao2016less,akata2015evaluation}, 
word embedding vectors trained from large text corpus using natural language processing (NLP) techniques
\cite{akata2015evaluation,frome2013devise,xian2016latent,zhang2015zero,al2016recovering}, 
co-occurrence statistics of hit-counts from search engine~\cite{rohrbach2010helps,mensink2014costa}, and WordNet hierarchy information of the labels~\cite{rohrbach2010helps,rohrbach2011evaluating,li2015zero}. 
These works demonstrated impressive results on several standard zero-shot datasets. 
However, majority research effort has concentrated on multi-class zero-shot classifications,
while the more challenging multi-label zero-shot learning problem has received  
very limited attention~\cite{mensink2014costa,zhang2016fast,lee2017multi}.

\begin{figure}[t]
\begin{center}
   \includegraphics[width=\linewidth]{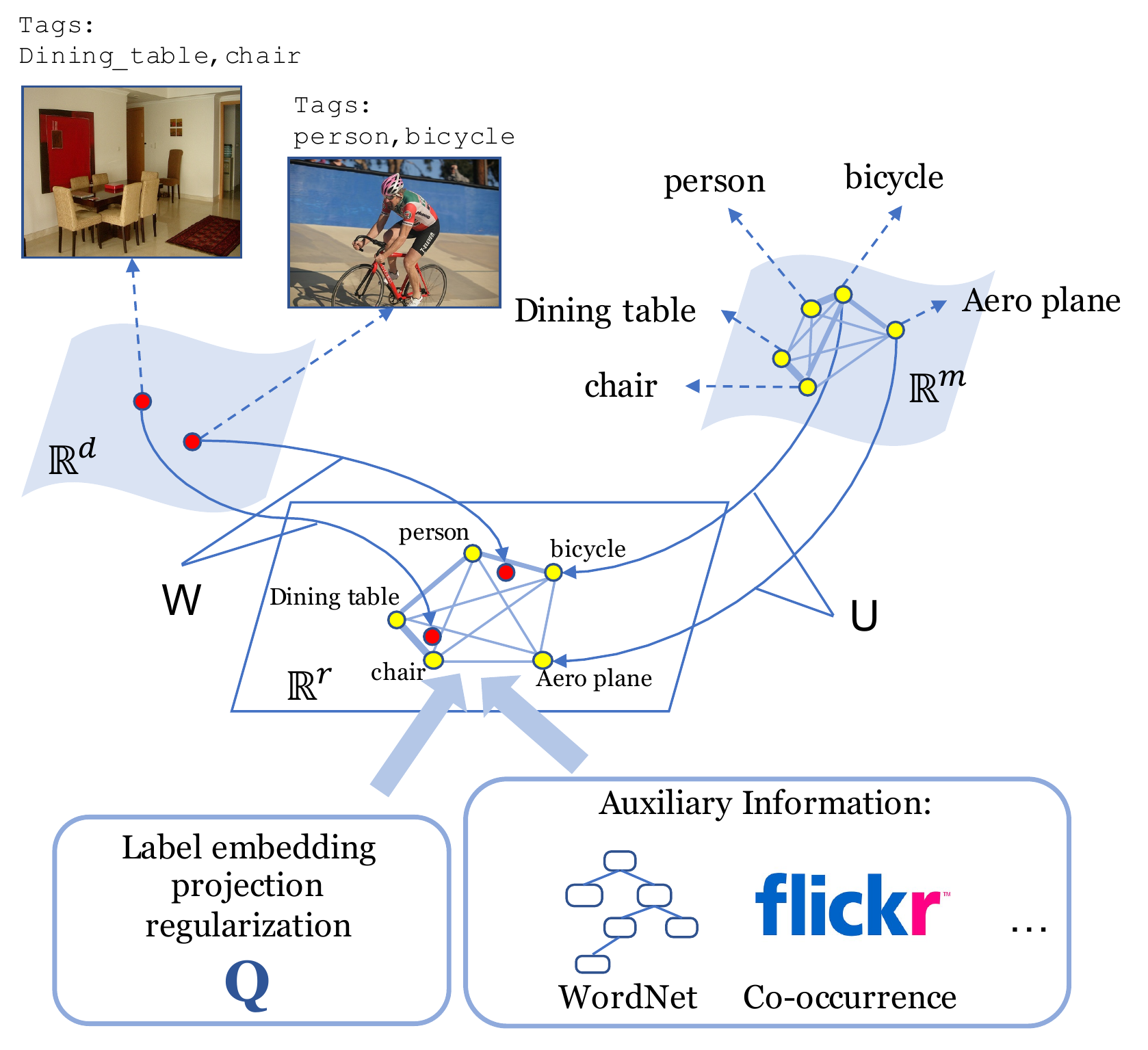}
\end{center}
\caption{Illustration of the proposed multi-label ZSL framework. 
Red dots represent images in their visual feature space $\mathbb{R}^d$. 
They are mapped into a semantic space $\mathbb{R}^r$ by a visual projection matrix $W$.
Yellow dots represent labels in the word embedding space $\mathbb{R}^m$ 
and they are mapped into the same $\mathbb{R}^r$ by a semantic projection matrix $U$.
The projection matrices are learnt 
under a max-margin multi-label learning framework
based on the matching scores of the images and labels 
in the projected semantic space.
Embedding regularization 
and auxiliary information are leveraged 
to facilitate the knowledge transfer from seen classes to unseen classes 
on the projected common semantic space.
}
\label{fig:TAEP}
\end{figure}

In this work we propose a novel transfer-aware label embedding projection method
to tackle multi-label zero-shot learning, as shown in Figure~\ref{fig:TAEP}. 
Label embeddings have been exploited in standard multi-label classification
to capture label relationships.
We exploit
the word embeddings~\cite{pennington2014glove} produced from large corpus with NLP techniques
as the initial semantic label embedding vectors. 
These semantic embedding vectors have the nice property of catching general similarities 
between any pair of label phrases/words, but may not be optimal for 
multi-label classification and information transfer across classes. 
Hence we project the label embedding vectors
into a low-dimensional semantic space 
in a transfer-aware manner 
to gain transferable label relationships 
by enforcing similarity between seen and unseen class labels
and separability across unseen labels.
We then simultaneously co-project the labeled seen class
instances into the same semantic space 
under a max-margin multi-label classification framework
to ensure the predictability of the embeddings. 
Moreover, we further incorporate auxiliary information 
to guide the label embedding projection
for suitable inter-label relationships.
To investigate the proposed approach,
we conduct ZSL experiments on  two standard multi-label image classification datasets,
the PASCAL VOC2007 and VOC2012.
The empirical results demonstrate the effectiveness of the proposed approach by 
comparing to a number of related ZSL methods. 

\section{Related Work}

\noindent{\bf Multi-label Classification}\quad
Multi-label classification is relevant in many application domains, 
where each data instance can be assigned into multiple classes.
Many multi-label learning works developed
in the literature have centered on exploiting 
the correlation/interdependency information 
between the multiple labels,
including 
the max-margin learning methods
with pairwise ranking loss~\cite{elisseeff2001kernel}, 
weighted approximate pairwise ranking loss (WARP)~\cite{weston2011wsabie},
and calibrated separation ranking loss (CSRL)~\cite{guo2011adaptive}.
Moreover, incomplete labels are frequently encountered 
in many multi-label applications due to noise or crowd-sourcing,
where only a subset of true labels are provided on some training instances.
Multi-label learning methods with missing labels
have largely depended on 
observed label correlations to overcome the label incompleteness
of the training data
\cite{BucakCVPR2011,yuetal13,yangeccv16}.
These methods however assumed that all the labels are 
at least observed on a subset of training data 
and they cannot handle the more challenging zero-shot learning setting
where some labels are completely missing from the training instances. 
%

\noindent{\bf Zero-shot Learning}\quad
There have been a significant number of works in multi-class zero-shot image classification, 
including the ones that  
explore different transferring embedding strategies
\cite{romera2015embarrassingly,frome2013devise,norouzi2013zero,xian2016latent} 
or different information sources
\cite{akata2015evaluation,mensink2014costa}. 
Many methods represent labels in a semantic
attribute space~\cite{lampert2009learning} or word embedding space~\cite{mikolov2013distributed,pennington2014glove}) to perform zero-shot learning by computing 
similarities between the instances and labels.
\cite{romera2015embarrassingly} proposed a simple approach to learn a projection matrix that maps 
image features into the attribute space, 
while~\cite{frome2013devise} used a CNN architecture followed by a transformation matrix to 
map images into the word embedding vector space. 
\cite{norouzi2013zero} also took advantage of CNNs 
but they expressed image embeddings as convex combinations of seen class embeddings. 
\cite{akata2015evaluation} considered learning a bilinear compatibility function for image features and output label embeddings. They evaluated attributes, word embedding vectors, as well as WordNet hierarchy and online text information, for producing label embeddings. 
In ~\cite{xian2016latent}, the authors proposed to use tensors as nonlinear latent embedding functions. 
\cite{li2015semi} learned the projection matrix by minimizing max-margin loss in a semi-supervised way. 
\cite{zhang2015zero} proposed to embed both image features and attribute signature of labels into a common semantic space which has the seen classes as bases. 
More recently, \cite{changpinyo2017predicting} proposed a method to generate visual examplars from semantic attributes, and then use them as optimized class prototypes for prediction on test instances. 
This work also projects both semantic and visual feature vectors into an intermediate space.
Nonetheless, all theses methods are designed for multi-class zero-shot learning problems.

Despite the many works above on multi-class ZSL, to the best of our knowledge, there has not been much work on multi-label ZSL with the following exceptions.
In ~\cite{fu2015multilabel}, the authors proposed to address multi-label zero-shot learning by mapping images into the semantic word space. 
However in testing phase it needs to consider all possible combinations of the outputs, which is the power set of unseen tags/classes. This prevents it from being applied on larger datasets. 
The authors of ~\cite{mensink2014costa} proposed to express unseen class classifiers 
as weighted sums of seen class classifiers, 
while the weights are estimated from different kinds of co-occurrence statistics. 
This approach however treats the unseen class classifiers separately,
without considering the correlations/dependencies among the classes.
\cite{zhang2016fast} proposed a fast zero-shot image tagging algorithm, 
which learns the principal direction of each image to separate tags into positive and negative ones. 
Their approach however uses fixed pre-given label embeddings which may not be the best for capturing 
useful class correlations between seen and unseen classes towards information transfer. 
More recently, 
in \cite{gaure2017probabilistic} the authors adopted a generative
probabilistic framework to leverage the co-occurrence statistics 
of the seen labels for multi-label zero-shot prediction. 
This method however heavily depends on the auxiliary resource 
for gaining quality label co-occurrence statistics.
\cite{lee2017multi} proposed to construct a knowledge graph based on WordNet hierarchy 
for modeling label relations, 
and then propagate confidence
scores from the seen to unseen labels through the graph. 
Its performance largely relies on the quality of the knowledge graph. 
By contrast, our proposed approach can project existing label embeddings into
a more suitable low-dimensional semantic space 
to automatically retrieve better label relations 
for knowledge transfer between seen and unseen classes,
while flexibly exploiting auxiliary information for additional help.

\section{Proposed Approach}
\subsection{Problem Definition and Notations}
We consider multi-label zero-shot learning in the following setting. 
Assume we have a set of $n$ labeled training images $D=(X,Y)$, 
where $X\in \mathbb{R}^{n\times d}$ denotes the $d$-dim visual features
extracted using CNNs
for the $n$ images, 
and $Y\in \{0,1\}^{n\times L^s}$
denotes the corresponding label indicator matrix across 
a set of seen classes, $\mathcal{S}=\{1,2,...,L^s\}$:
``1'' indicates the presence of the corresponding label (i.e., positive labels)
and ``0'' indicates the absence of the corresponding label (i.e., negative labels).
For multi-label classification, each row of $Y$ can have multiple 
``1'' values.
Moreover, we also assume there are a set of $L^u$
unseen classes, $\mathcal{U}=\{L^s+1,...,L\}$ such that $L=L^s+L^u$, 
and the labels
for the unseen classes are completely missing in our labeled training data.
In addition, we assume the word embeddings of the seen classes and unseen classes 
are both given: $M = [{M^s}; {M^u}]\in\mathbb{R}^{L\times m}$, 
where $M^s\in \mathbb{R}^{L^s\times m}$ are the seen class embeddings, 
$M^u\in \mathbb{R}^{L^u\times m}$ are the unseen class embeddings,
and their concatenation $M$ is for all the classes. 
We aim to learn a multi-label prediction model from the training data
that allows us to perform multi-label classification on the unseen classes.

We use the following general notations in the presentation below.
For any matrix, e.g., $X$, we use $X_i$ to denote its $i$-th row vector.
We use $\|\cdot\|_{F}$ to denote the Frobenius norm of a matrix
and use tr($\cdot$) to denote the trace of a matrix.
For $Y_i$, we use $\bar{Y}_i$ to denote its complement 
such that  $\bar{Y}_i=1-Y_i$. We also reuse the notation $Y_i$ to 
denote a set of indices of its non-zero values within proper contexts.
We use $\|\cdot\|$ to denote the Euclidean norm
and denote the rectified operator as $[\cdot]_+=\max(\cdot,0)$.
We use  {\bf 1} to denote a column vector of all 1s, assuming its size can be 
determined in the context,
and use $I$ to denote an identity matrix.
We use ${\bf 0}_{a,b}$ to denote a $a\times b$ matrix with all 0s
and use ${\bf 1}_{a,b}$ to denote a $a\times b$ matrix with all 1s.

\subsection{Max-margin Multi-label Learning with Semantic Embedding Projection}

Instead of entirely relying on the pre-given label embeddings in $M$
obtained from word embeddings
to facilitate cross-class information adaptation, 
we propose to co-project the input image visual features and the label embeddings
into a more suitable common low-dimensional semantic space
such that the similarity matching scores of each image with its positive labels
in this semantic space will be higher than that with its negative labels.
Specifically, 
we want to learn a projection function $\theta:\mathbb{R}^d\rightarrow \mathbb{R}^r$ 
that maps an instance $X_i$ from the visual feature space $\mathbb{R}^d$ into a semantic space $\mathbb{R}^r$;
assuming a linear projection we have $\theta(X_i)=X_iW$, 
where $W$ is a $d\times r$ projection matrix. 
Simultaneously, we learn another 
linear projection function $\phi:\mathbb{R}^m\rightarrow\mathbb{R}^r$ 
such that $\phi(M_c) = M_cU$,
where $U$ is a $m\times r$ projection matrix, 
which maps a class $c$ from the original word embedding space $\mathbb{R}^m$ 
into the same semantic space $\mathbb{R}^r$.
Then the similarity matching score between an instance $X_i$ and 
the $c$-th class label can be computed as the inner product
of their project representations in the common semantic space:
\begin{equation}
\label{eq:dot_prod}
F(i,c) = \theta(X_i)\phi(M_c)^\top = X_iW\, U^\top M_c^\top
\end{equation}

To encode the assumption that the similarity score 
$F(i,c)$ between an instance $X_i$ and any of its positive label $c\in Y_i$
should be higher than the similarity score
$F(i,\bar{c})$ between instance $X_i$ and any of its negative label $\bar{c}\in\bar{Y}_i$,
i.e., $F(i,c)\succ F(i,\bar{c})$,
we formulate the projection learning problem within a max-margin multi-label
learning framework:
\begin{equation}
\min_{W, U: U^\top U = I} \quad \sum_{i=1}^n\mathcal{L}(W,U;X_i,Y_i) + \mathcal{R}(W) 
\label{framework}
\end{equation}
where $\mathcal{L}(\cdot)$ denotes a max-margin ranking loss 
and $\mathcal{R}(W)$ is a model regularization term.  
In this work we adopt a calibrated separation ranking loss: 
\begin{equation}
\label{eq:CSRL}
\begin{aligned}
\mathcal{L}(W,U;X_i,Y_i) \!=\! 
\left\{\!\!
\begin{array}{l}
\max_{c\in Y_i}\big[1 \!+\! F(i,0)\!-\!F(i,c)\big]_+ \\[1.5ex]
\!+\! \max_{\bar{c}\in\bar{Y_i}}\big[1 \!+\! F(i,\bar{c}) \!-\! F(i,0)\big]_+
\end{array}
\!\!\!\right\}
\end{aligned}
\end{equation}
where $F(i,0)=X_i W_0$ can be considered
as the matching score for an auxiliary class 0,
which produces 
a separation threshold score 
on the $i$-th instance 
such that the scores for positive labels
should be higher than it and the scores for negative labels should be lower than it,
i.e., $F(i,c)\succ F(i,0) \succ F(i,\bar{c})$,
to minimize the loss.

We assume the project matrix $U$ has orthogonal columns to maintain
a succinct label embedding projection. 
For the regularization term over $W$,
we consider a Frobenius norm regularizer, 
$\mathcal{R}(W) = \frac{\beta}{2}\left(\|W\|^2_{\mathrm{F}} 
+ \|W_0\|^2\right)$,
where 
$W_0$ can be considered as an auxiliary column to $W$,
and $\beta$ is a trade-off weight parameter.

\subsection{Transfer-Aware Label Embedding Projection}

Employing the ranking loss to minimize classification error on seen classes 
can ensure the predictability of the projected label embedding. 
However for ZSL our goal is to predict labels from the unseen classes. 
This requires a label embedding representation
that can encode suitable inter-class label relations
to facilitate information transfer from seen to the unseen classes
such that the similarity score $F(i,c)$ can 
well reflect the relative prediction scores on an unseen class $c$
under the learned model parameters $W$ and $U$.
Our intuition is that 
classification or ranking on the target unseen class labels 
would be easier if they are well separated in the projected embedding space
and knowledge transfer would be easier if unseen classes and seen classes 
have high similarities in the projected label embedding space.
We hence propose to guide the label embedding projection learning
by encoding this intuition through a transfer-aware regularization objective $\mathcal{H}(U)$
such that:
\begin{align}
\mathcal{H}(U) 
= & 
\frac{\gamma}{2L^u(L^u-1)}
\sum_{i,j\in\mathcal{U}, i\not=j}
M_iUU^\top M_j^\top -
\nonumber\\
&\frac{\gamma}{2L^sL^u}\sum_{i\in\mathcal{S},j\in\mathcal{U}} M_iUU^\top M_j^\top
\nonumber
\end{align}
which can be equivalently expressed in a more compact form:
\begin{align}
\mathcal{H}(U) 
&= \frac{\gamma}{2}\mathrm{tr}\left(U^\top M^\top QMU\right)
\label{projreg}
\end{align}
where $\gamma$ is a balance parameter for $\mathcal{H}(\cdot)$, 
and 
\begin{equation}
Q = 
\begin{bmatrix}
\boldsymbol{0}_{L^s,L^s} & \frac{-1}{2L^sL^u}\boldsymbol{1}_{L^s,L^u} \\
\frac{-1}{2L^sL^u}\boldsymbol{1}_{L^u,L^s} & \frac{1}{L^u(L^u-1)}(\boldsymbol{1}_{L^u,L^u}-I_{L^u})
\end{bmatrix}
\end{equation}
Here we use the inner product of a pair projected label embedding vectors as the 
similarity value for the corresponding pair of classes, and aim to maximize the similarities across seen
and unseen classes and minimize the similarities between unseen classes.
By incorporating this regularization objective into the framework in Eq.(\ref{framework}),
we obtain the following Transfer-Aware max-margin Embedding Projection (TAEP) learning problem:
\begin{align}
\!\!\!
\min_{\substack{W,W_0,\xi,\eta,\\ U:\;  U^\top U=I}}
& 
\boldsymbol{1^\top\xi} + \boldsymbol{1^\top\eta} 
+ \frac{\beta}{2}(\|W\|^2_{\mathrm{F}}\!+\!\|W_0\|^2) 
+ \mathcal{H}(U) 
\label{eq:final}\\
\text{s.t.} \quad
& F(i,c) - F(i,0) \geq 1-\boldsymbol{\xi}_i , \forall c\in Y_i , \forall i;
\;\boldsymbol{\xi}\geq 0;  
\nonumber\\
& F(i,0) - F(i,\bar{c}) \geq 1-\boldsymbol{\eta}_i , \forall \bar{c}\in \bar{Y_i} , \forall i; 
\; \boldsymbol{\eta}\geq 0 \nonumber
\end{align}
The objective learns $W$ and $U$ by enforcing positive labels to rank higher than negative labels, 
while incorporating the regularization term 
$\mathcal{H}(U)$ 
to refine the label embedding structure 
in the semantic space. 
$\mathcal{H}(U)$ 
can help produce better inter-class relationship structure for cross-class knowledge transfer.
The regularization form 
$\mathcal{H}(U)$ 
also has a nice property ---  
it allows a closed-form solution for $U$ to be derived 
and hence simplifies the training procedure.


Note after learning the projection matrices $W$ and $U$,
it will be straightforward to rank all unseen labels for instance $i$ 
based on the prediction scores $F(i,c)$ for all $c\in\mathcal{U}$.

\subsection{Integration of Auxiliary Information }

In addition to explicit word embeddings, 
similarity information about the class labels 
can be derived from
some external resources. 
We propose to leverage such auxiliary information to further improve label embedding projection.

In general, we can 
assume there is some auxiliary source in terms of a similarity matrix $R$ over the seen and unseen labels; i.e., $R_{ij}, i,j\in\{1,2,...,L\}$ defines the similarity between a label pair $(i,j)$. 
Then $Q^{\mathcal{A}} = I - D^{-1/2}RD^{-1/2}$, where $D=diag(R\boldsymbol{1})$, is the normalized Lapalacian matrix of $R$. 
We use a manifold regularization term to enforce the 
projected label embeddings to be better aligned with the inter-class affinity $R$:
\begin{equation}
\mathcal{A}(U) = \frac{\lambda}{2}\mathrm{tr}\left(U^\top M^\top Q^{\mathcal{A}}MU\right)
\end{equation}
where $\lambda$ is a balance parameter for $\mathcal{A}(\cdot)$. 
This regularization form has the following advantages. 
First, it can be conveniently integrated into the learning framework
in Eq.(\ref{eq:final}) by simply updating the regularization function $\mathcal{H}(U)$ to:
\begin{align}
\mathcal{H}(U) 
= \frac{\gamma}{2}\mathrm{tr}\left(U^\top M^\top (Q + \frac{\lambda}{\gamma} Q^{\mathcal{A}} )MU\right)
\label{projreg2}
\end{align}
Second, it is convenient to exploit different auxiliary resources by 
simply replacing $R$ (or $Q^{\mathcal{A}}$) with the one computed from the specific resource.
In this work we study two different auxiliary information resources,
WordNet~\cite{miller1995wordnet} hierarchy and web co-occurrence statistics.

\vskip .1in
\noindent{\bf WordNet:}\quad 
WordNet~\cite{miller1995wordnet} 
is a large lexical database of English. Words are grouped into a hierarchical tree structure based on their semantic meanings. 
Since words are organized based on ontology, 
their semantic relationships can be reflected by their connection paths.
We find the shortest path between any two words based on ``is-a'' taxonomy, 
and then define the similarity between two labels $i$ and $j$ as 
the reciprocal of the path length between the corresponding words, i.e., $R_{ij} = \frac{1}{path\_len(i,j)+1}$.

\vskip .1in
\noindent{\bf Co-occurrence statistics:}\quad 
Many researchers have exploited the usage of online data, for example Hit-Count, to compute similarity between labels ~\cite{rohrbach2010helps,mensink2014costa}. The Hit-Count $HC(i,j)$ denotes how many times in total $i$ and $j$ appear together in the auxiliary source -- for example, the number of records returned by a search engine. It is the co-occurrence statistics of $i$ and $j$ in the scale of the entire World Wide Web. 
Following previous works, we use the Flickr Image Hit Count 
to compute the dice-coefficient as similarity between two labels, i.e., $R_{ij} = \frac{HC(i,j)}{HC(i)+HC(j)} $.

\subsection{Dual Formulation and Learning Algorithm}

With the orthogonal constraint on $U$
and the appearance of $U$ in both the objective function
and the linear inequality constraints,
it is difficult to perform learning directly on Eq.(\ref{eq:final}).
We hence deploy the standard Lagrangian dual formulation of the max-margin
learning problem for fixed $U$.
This leads to the following equivalent dual formulation of Eq.(\ref{eq:final}):
\begin{align}
\min_{U:U^\top U=I} 
&
\max_{\Psi}\; 
\mathrm{tr}\left(\Psi^\top(2Y\!-\!\boldsymbol{11}^\top)\right) 
\!+\! \frac{\gamma}{2}(U^\top M^\top QMU) 
\nonumber\\
&\!\!\!\!\!\!\!\!\!
- \frac{1}{2\beta}\mathrm{tr}\Big(\Psi^\top XX^\top \Psi 
\big(M^s 
 UU^\top{M^s}^\top \!+\!\boldsymbol{11}^\top\big)\Big) 
\label{eq:dual}
\\
\text{s.t.} \quad
& \Psi_i\mathrm{diag}(Y_i)\geq 0, \quad \Psi_iY_i^\top\leq 1, \;\forall i;\nonumber\\
& \Psi_i\mathrm{diag}(Y_i-1)\geq 0, \quad \Psi_i(Y_i-1)^\top\leq 1, \; \forall i
\nonumber
\end{align}
where the primal $W$ and $W_0$ can be recovered from the dual variables
$\Psi$ by
$W = \frac{1}{\beta}X^\top\Psi M^sU$ and 
$W_0 = \frac{-1}{\beta}X^\top\Psi {\bf 1}$.

One nice property about the dual formulation in Eq.(\ref{eq:dual})
is that it allows a convenient closed-form solution for $U$.
To solve this min-max optimization problem, 
we develop an iterative alternating optimization algorithm to perform training.
We start from an infeasible initialization point 
by setting both $U$ and $\Psi$ as zeros. 
Then in each iteration, we perform the following two steps,
which will quickly move into the feasible region 
after one iteration.

\paragraph{\bf Step 1:} 
Given the current fixed $U$,  
the inner maximization over $\Psi$ is a linear constrained 
convex quadratic programming. 
Though we can solve it directly using a quadratic solver, 
it subjects to a scalability problem--
the Hessian matrix over $\Psi$ will be very large whenever 
the data size $n$ or the label size $L^s$ is large.
Hence we adopt a coordinate descent method 
to iteratively update each row of $\Psi$ given other rows fixed,
since the constraints over each row of $\Psi$ can be separated.
The maximization over the $i$-th row $\Psi_i$ can be equivalently
written as the following simple quadratic minimization problem:
\begin{align}
\label{eq:quadratic}
\boldsymbol{z}^* = 
 \argmin_{\boldsymbol{z}}& \quad \frac{1}{2}\boldsymbol{z}^\top H\boldsymbol{z} + \boldsymbol{f}^\top\boldsymbol{z} \\
\text{s.t.}
& \quad \mathrm{diag}(Y_i)\boldsymbol{z}\geq 0, \quad \mathrm{diag}(Y_i-1)\boldsymbol{z}\geq 0, \nonumber\\
& \quad Y_i\boldsymbol{z}\leq 1, \quad (Y_i-1)\boldsymbol{z}\leq 1 \nonumber
\end{align}
where $H=\frac{1}{\beta}X_i{X_i}^\top(M^sUU^\top {M^s}^\top + \boldsymbol{11}^\top)$ and
$\boldsymbol{f} = \boldsymbol{1} - 2Y_i^\top + \frac{1}{\beta}(M^sUU^\top {M^s}^\top + \boldsymbol{11}^\top)\Psi^\top X X_i^\top$. 
After obtaining the optimal solution $\boldsymbol{z}^*$, 
we can update $\Psi$ with $\Psi \gets \Psi + \boldsymbol{1}_i{\boldsymbol{z}^*}^\top$, where $\boldsymbol{1}_i$ denote a one-hot vector with a single 1 in its $i$-th entry
and 0s in all other entries.

\vskip .1in
\noindent{\bf Step 2:}\quad 
After updating each row in $\Psi$, 
we fix the value $\Psi$ and perform minimization over $U$. 
By taking a negative sign from Eq.(\ref{eq:dual}), 
we have the following maximization problem:
\begin{equation}
\label{eq:min_u}
\max_{U:U^\top U=I}\mathrm{tr}\Big(U^\top\big(\frac{1}{2\beta}{M^s}^\top\Psi^\top XX^\top\Psi M^s - \frac{\gamma}{2}M^\top QM\big)U\Big)
\end{equation}
which has a closed-form solution.
Let $S=\frac{1}{2\beta}{M^s}^\top\Psi^\top XX^\top\Psi M^s - \frac{\gamma}{2}M^\top QM$. 
Then the solution for $U$ is the top-r eigenvectors of $S$.

\begin{table*}[!ht]
\setlength{\tabcolsep}{5pt}
\begin{center}
\caption{Average comparison results (\%) over five runs on zero-shot multi-label image tagging.
Smaller values indicate better results in terms of Hamming loss,
while larger values indicate better results in terms of the remaining measures.
}
\label{tbl:zero}
\begin{tabular}{|l|c|c|c|c||c|c|c|c|c|}
\hline 
\multirow{2}{*}{Methods} & \multicolumn{4}{c||}{VOC2007} & \multicolumn{4}{c|}{VOC2012} \\ \cline{2-9} 
& MiAP   & micro-F1 & macro-F1 & Hamm. & MiAP & micro-F1 & macro-F1 & Hamm.\\ 
\hline \hline 
ConSE  & 49.98& 30.80 & 27.57 & 28.12 & 49.95 & 33.48 & 28.83 & 27.13 \\   
\hline 
LatEm-M & 52.45& 35.32 & 36.69 & 26.28 & 51.44& 35.74 & 36.33 & 26.21 \\  
\hline 
DMP      & 53.52 & 36.70 & 40.44 & 25.72 & 52.92 & 35.73 & 41.04 & 26.12 \\
\hline
Fast0Tag & 52.39 & 35.01 & 36.76 & 26.53 & 52.29 	& 34.23 & 35.38 & 26.41 \\
\hline \hline
TAEP     & 57.42 & 38.48 & 42.33 & 24.98 & 54.39	& 37.63 	& 41.58 & 25.25\\ 
\hline  
TAEP-C    & {\bf 59.22} & {\bf 39.84} & {\bf 43.77} & {\bf 24.01} & {\bf 57.13} & {\bf 39.30} & {\bf 42.97} & {\bf 24.27} \\  
\hline
TAEP-H    & 57.62 & 38.95 & 43.29 & 24.46  & 56.10 & 38.89 & 42.23 & 24.44 \\
\hline
\end{tabular} 
\end{center}
\end{table*}
\section{Experiments}
To investigate the empirical performance of the proposed method,
we conducted experiments on two standard multi-label image classification datasets
to test its performance on multi-label zero-shot classification
and generalized multi-label zero-shot classification. 
\subsection{Experimental Setting}
\label{sec:setting}
\paragraph{Datasets}
In our experiments we used two standard multi-label datasets: 
The PASCAL VOC2007 dataset 
and VOC2012 dataset. 
The PASCAL VOC2007 dataset contains 20 visual object classes. There are 9963 images in total, 5011 for training and 4952 for testing. 
The VOC2012 dataset 
contains 5717 and 5823 images from 20 classes for training and validation. 
We used the validation set for test evaluation. 

\vskip .1in
\noindent {\bf Detailed settings}\quad
For each image, we used 
VGG19~\cite{simonyan2014very} pre-trained on ImageNet
to extract the 4096-dim 
visual features. 
For the label embeddings, we used the 300-dim word embedding vectors pre-trained 
by GloVe~\cite{pennington2014glove}. 
All image feature vectors and word embedding vectors are $\mathit{l}_2$ normalized.
To determine the hyper-parameters, we further split the seen classes into two disjoint subsets with equal number of classes for training and validation. 
We train the model on the training set and choose hyper-parameters based on the test performance on the validation set. 
For the proposed model, we choose $\beta$, $\gamma$ and $\lambda$ from 
$\beta\in\{1, 2,..., 10\}$ and $\gamma,\lambda\in\{0.01, 0.1, 1, 10\}$ respectively.
After parameter selection, the training and validation data are put back together 
to train the model for the final evaluation on unseen test data.

\vskip .1in
\noindent\textbf{Evaluation metric}\quad
We used four different multi-label evaluation metrics: MiAP, micro-F1, macro-F1 and Hamming loss.
The Mean image Average Precision (MiAP)~\cite{li2016socializing} 
measures how well are the labels ranked on a given image based on the prediction scores. 
The other three standard evaluation metrics for multi-label classification 
measure how well the predicted labels match with the ground truth labels on the test data.
%

\subsection{Multi-label Zero-shot Learning Results}
\label{sec:VOC}
\paragraph{Comparison methods}
We compared the proposed method with 
four related multi-label ZSL methods, {\em ConSE, LatEm-M}, {\em DMP} and {\em Fast0Tag}, 
which also adopted the visual-semantic projection strategy. 
The first two methods are the multi-label adaptations of 
two standard ZSL approaches,
the convex combination of semantic embedding (ConSE)~\cite{norouzi2013zero} 
and the latent embedding (LatEm) method~\cite{xian2016latent}.  
For LatEm, we adopted a multi-label ranking objective to replace the original one of LatEm 
and denote this variant as Latent Embedding Multi-label method ({\em LatEm-M}). 
The direct multi-label zero-shot prediction method (DMP)~\cite{fu2015multilabel} 
and the fast tagging method (Fast0Tag)~\cite{zhang2016fast} 
are specifically developed for mulit-label zero-shot learning.
For our proposed transfer-aware max-margin embedding projection ({\em TAEP}) method, 
we also provide comparisons for two TAEP variants with 
different types of auxiliary information: 
{\em TAEP-H} uses WordNet Hierarchy as auxiliary information, 
and {\em TAEP-C} uses Flickr Image Hit-Count as auxiliary information.
\\[2ex]

\noindent{\bf Zero-shot multi-label learning results. }\quad 
We divided the datasets into two subsets of equal number of classes, and then use them as seen and unseen classes 
respectively. All methods use seen class instances in the training set to train their models 
and make predictions on the unseen class instances in test set. 
We selected the hyper-parameters for the comparison methods based on grid search. 
With selected fixed parameters,
for each approach we repeated 5 runs and reported its mean performance in Table~\ref{tbl:zero}. 
We can see the direct multi-label prediction method, DMP, outperforms both ConSE and LatEm-M
on the two datasets in terms of almost all measures.
This shows that the specialized multi-label ZSL method, DMP, does have advantage over extended multi-class ZSL methods.
Fast0Tag is a bit less effective than DMP, but
still consistently outperforms ConSE. 
The proposed TAEP on the other hand consistently outperforms 
all the four comparison methods across all measures 
and with notable improvements on both datasets.
By integrating auxiliary information, the proposed TAEP-C and TAEP-H 
further improve the performance of the proposed model TAEP,
while TAEP-C achieves the best results in terms of all measures. 
These results verified the efficacy of the proposed model. 
They also demonstrated the usefulness of auxiliary information and 
validated the effective information integration mechanism of our proposed model.
\\[2ex]
\begin{table*}[!htbp]
\setlength{\tabcolsep}{5pt}
\begin{center}
\caption{Average comparison results (\%) on generalized multi-label zero-shot Learning.
Smaller values indicate better results in terms of Hamming loss,
while larger values indicate better results in terms of the remaining measures.}
\label{tbl:gzsl}
\begin{tabular}{|l|c|c|c|c||c|c|c|c|c|}
\hline 
\multirow{2}{*}{Methods} & \multicolumn{4}{c||}{VOC2007} & \multicolumn{4}{c|}{VOC2012} \\ \cline{2-9} 
& MiAP     & micro-F1 & macro-F1 & Hamm. & MiAP& micro-F1 & macro-F1 & Hamm.\\ 
\hline \hline 
ConSE  & 64.10& 42.11 & 32.29 & 12.78 & 62.85 & 41.17 & 35.72 & 13.04 \\  
\hline
LatEm-M & 66.46 & 43.11 & 32.37 & 12.56 & 63.06 & 39.95 & 32.35 & 13.31 \\  
\hline
DMP      & 67.79 & 43.97 & 34.13 & 12.37 & 64.24 & 41.29 & 32.39 & 13.02 \\
\hline
Fast0Tag & 67.34 & 43.54 & 33.31 & 12.49 & 64.63 	& 41.28 & 32.46 & 12.97 \\
\hline \hline
TAEP     & 68.16 & 43.61 & 35.29 & 12.01 & 64.67 	& 40.60 & 34.07 & 12.75 \\ 
\hline  
TAEP-C   & {\bf 69.87} & {\bf 44.75} & {\bf 35.62} & {\bf 11.98} & {\bf 65.33} & {\bf 42.10} & {\bf 36.74} & {\bf 12.53} \\  
\hline
TAEP-H   & 69.74 & 44.55 & 35.56 & 12.00 & 65.10 & 41.39 & 35.95 & 12.94 \\
\hline
\end{tabular} 
\end{center} 
\end{table*}
%
\noindent{\bf Generalized multi-label zero-shot learning results.}\quad 
Although zero-shot learning has often been evaluated only on the unseen classes in the literature,
it is natural to evaluate multi-label zero-shot learning on all the classes,
which is referred to as generalized multi-label zero-shot learning.  
Hence we conducted experiments to test the generalized zero-shot classification performance of the comparison methods.
Each method is still trained on the same seen classes $\mathcal{S}$, 
but the test set now contains all the seen and unseen labels, i.e., $\mathcal{S}\cup\mathcal{U}$.
The average comparison results  on the two datasets are reported in 
Table~\ref{tbl:gzsl}. 
We can see that the two specialized multi-label zero-shot learning methods, DMP and Fast0Tag, 
outperform the adapted methods ConSE and LatEm-M in terms of most measures on both VOC2007 and VOC2012, 
while TAEP achieves competitive performances with them.
By further incorporating the auxiliary information,
the proposed methods, TAEP-C and TAEP-H, not only consistently outperform
all the three comparison methods on both datasets in terms of all the evaluation metrics,
they also consistently outperform the base model TAEP. 
TAEP-C again produced the best results in most cases.
These results suggest our proposed model provides an effective framework
on learning transfer-aware label embeddings 
for generalized multi-label zero-shot learning,
and it also provides the effective mechanism on incorporating 
free auxiliary information. 
\begin{figure}[t!]
\begin{center}
\includegraphics[width=0.45\linewidth]{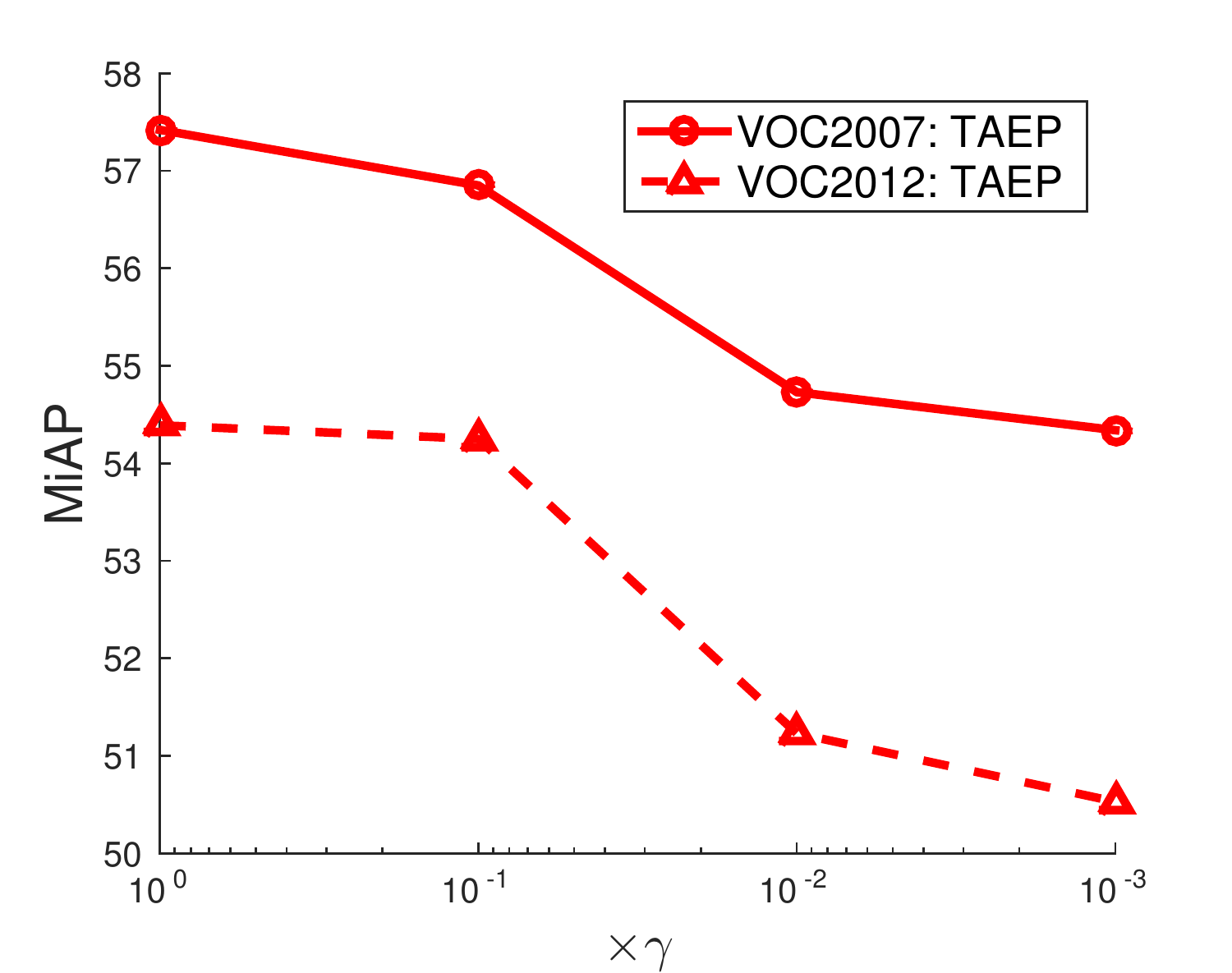}
\includegraphics[width=0.45\linewidth]{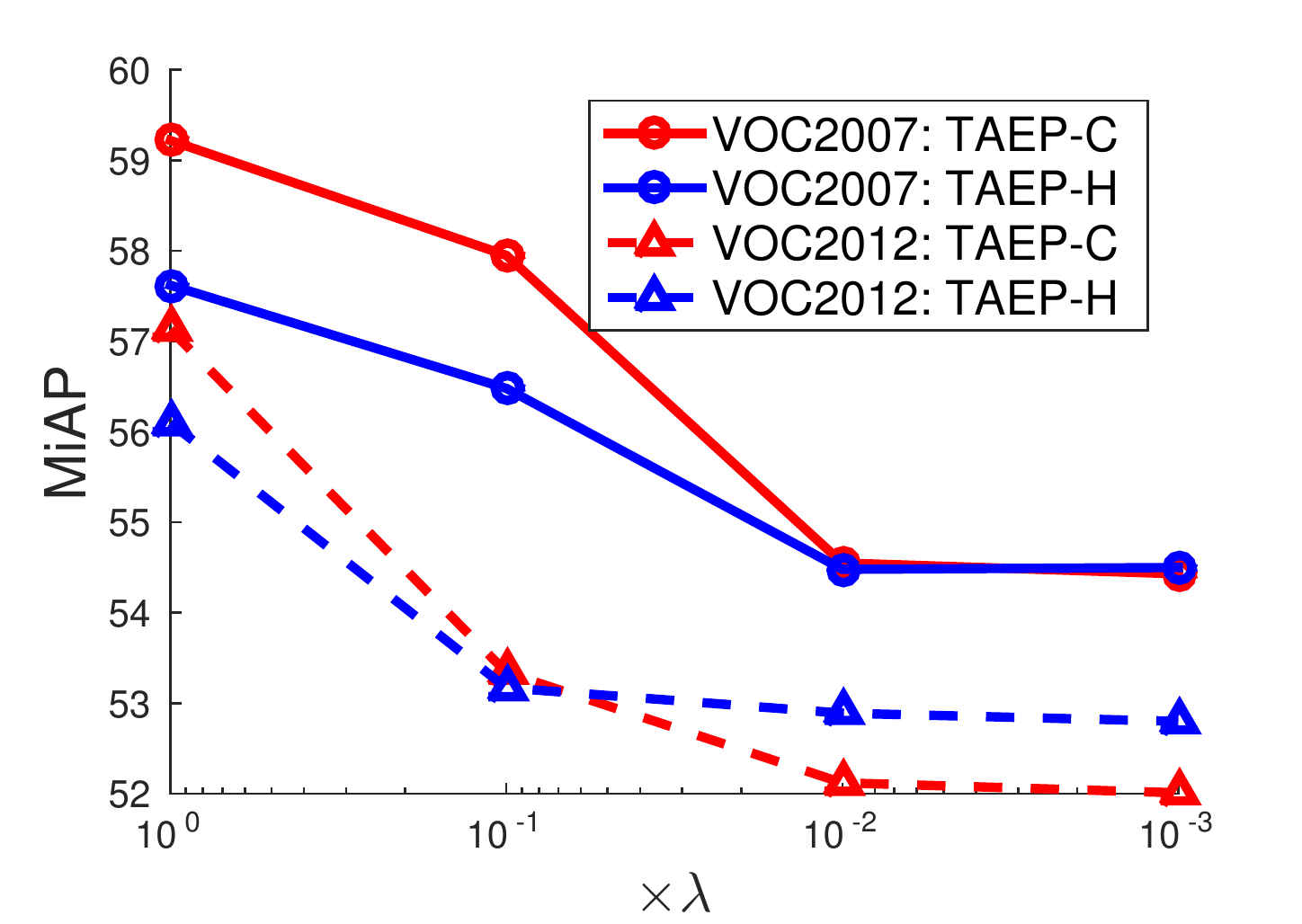}
\end{center}
\vskip -.1in
\caption{Impacts of the $\mathcal{H}(U)$ term 
and the auxiliary information. 
Note x-axis shows relative scaling factors on $\gamma$ or $\lambda$.
By gradually diminishing the regularization term (via $\gamma$, on the left)
or the auxiliary information (via $\lambda$, on the right), the performance drops. 
}
\label{fig:gamma}
\end{figure}


\subsection{Impact of Label Embedding Regularization}
In this section we study the impact of label embedding projection regularization term $\mathcal{H}(U)$,
i.e., the transfer-aware part of the proposed model.
For TAEP, 
we firstly set the parameters to the same values, $\gamma_0$, as those that generate Table \ref{tbl:zero}, 
and then reduce
$\gamma$ by a factor of 10 each time to repeat the experiments. 
That is, we try $\gamma$=$\gamma_0\times$\{$10^0, 10^{-1}, 10^{-2}, 10^{-3}$\}. 
Since $\gamma$ is the weight for the regularization term $\mathcal{H}(U)$, 
by doing this we are actually reducing 
the contribution of the embedding 
projection regularization term. 
The results in terms of MiAP are presented in Figure \ref{fig:gamma}.
Similarly, we also tested the impact of auxiliary information through the regularization 
term $\mathcal{H}(U)$ for TAEP-H and TAEP-C 
by reducing $\lambda$ by factors of $\{10^0,10^{-1},10^{-2},10^{-3}\}$.
From Figure~\ref{fig:gamma} we can see that, as $\gamma$ decreases, the performance of TAEP 
decreases on both datasets. 
This suggests that the label embedding projection regularization term $\mathcal{H}(U)$ 
is a necessary and useful component.
By regularizing the label embeddings to induce better inter-label relationships,
the cross-class information transfer can be facilitated in zero-shot learning. 
Similarly, we also observe that when $\lambda$ decreases, the performance of TAEP-C and TAEP-H
decreases as well on both datasets.
This again verifies the usefulness of auxiliary information
and the effectiveness of auxiliary integration mechanism of the proposed transfer-aware embedding projection method.
\section{Conclusion}
In this paper we proposed a transfer-aware label embedding approach for 
multi-label zero-shot image classification. 
This approach projects both images and labels into the same semantic space to 
rank the similarity scores of the images with positive and negative labels 
under a max-margin learning framework, 
while guiding the label embedding projection 
with a transfer-aware regularization objective 
to achieve a suitable inter-label relations 
for information adaptation. 
The regularization framework also allows convenient incorporations of auxiliary information. 
We conducted experiments to compare our approach with a few related ZSL methods
on multi-label image classification tasks. 
The results demonstrated the efficacy of the proposed approach.

\bibliographystyle{named}
\bibliography{paperbib}

\end{document}